\pdfoutput=1
\documentclass[11pt]{article}

\usepackage[]{acl}

\usepackage{times}
\usepackage{latexsym}

\usepackage[T1]{fontenc}
\usepackage[utf8]{inputenc}

\usepackage{microtype}

\usepackage{amsfonts}
\usepackage{amsmath}
\usepackage{algorithm, algpseudocode}
\usepackage{graphicx}
\usepackage{multirow}
\usepackage{array}
\usepackage{booktabs}
\usepackage{pifont}
\newcommand{\cmark}{\ding{51}}%
\newcommand{\xmark}{\ding{55}}%

\newcommand{\red}{\textcolor{red}}
\newcommand{\blue}{\textcolor{blue}}

\newcommand{\orange}{\textcolor{orange}}
\usepackage{verbatim}
\usepackage{stmaryrd}
\usepackage{trimclip}

\makeatletter
\DeclareRobustCommand{\shortto}{%
  \mathrel{\mathpalette\short@to\relax}%
}

\newcommand{\short@to}[2]{%
  \mkern2mu
  \clipbox{{.5\width} 0 0 0}{$\m@th#1\vphantom{+}{\shortrightarrow}$}%
  }
\makeatother

\title{Continual Sequence Generation with Adaptive Compositional Modules}

\author{Yanzhe Zhang \\
  Georgia Institute of Technology \\
  \texttt{z\_yanzhe@gatech.edu} \\\And
  Xuezhi Wang \\
  Google \\
  \texttt{xuezhiw@google.com} \\\And
  Diyi Yang \\
  Georgia Institute of Technology \\
  \texttt{dyang888@gatech.edu} \\}

\begin{document}
\maketitle
\begin{abstract}
Continual learning is essential for real-world deployment when there is a need to quickly adapt the model to new tasks without forgetting knowledge of old tasks. Existing work on continual sequence generation either always reuses existing parameters to learn new tasks, which is vulnerable to catastrophic forgetting on dissimilar tasks, or blindly adds new parameters for every new task, which could prevent knowledge sharing between similar tasks. To get the best of both worlds, in this work, we propose continual sequence generation with adaptive compositional modules to adaptively add modules in transformer architectures and compose both old and new modules for new tasks. We also incorporate pseudo experience replay to facilitate knowledge transfer in those shared modules. Experiment results on various sequences of generation tasks show that our framework can adaptively add modules or reuse modules based on task similarity, outperforming state-of-the-art baselines in terms of both performance and parameter efficiency.
We make our code public at \url{https://github.com/GT-SALT/Adaptive-Compositional-Modules}.
\end{abstract}

\section{Introduction}

Current state-of-the-art language generation models can achieve great performance on a wide range of sequence generation tasks \citep{radford2019language, lewis-etal-2020-bart} with a static data distribution. However, real-world scenarios are often changing which requires the model to learn with dynamic data distributions.
In such cases of data distributions shift, current generation models often suffer from \emph{catastrophic forgetting} \citep{sun2019lamol}: models completely and abruptly forget previously learned information upon learning new information.
% a model's knowledge on previous tasks is severely forgotten when its parameters are adapted towards learning knowledge from new tasks. 
% While 
Continual learning (CL) \citep{ring1998child, thrun1998lifelong} has been introduced to improve model's ability to learn tasks in a stream by mitigating forgetting and facilitating knowledge transfer \citep{lopez2017gradient}, however, continual sequence generation is relatively under-investigated.

\begin{figure}[]
\begin{center}
\includegraphics[width=1.0\linewidth]{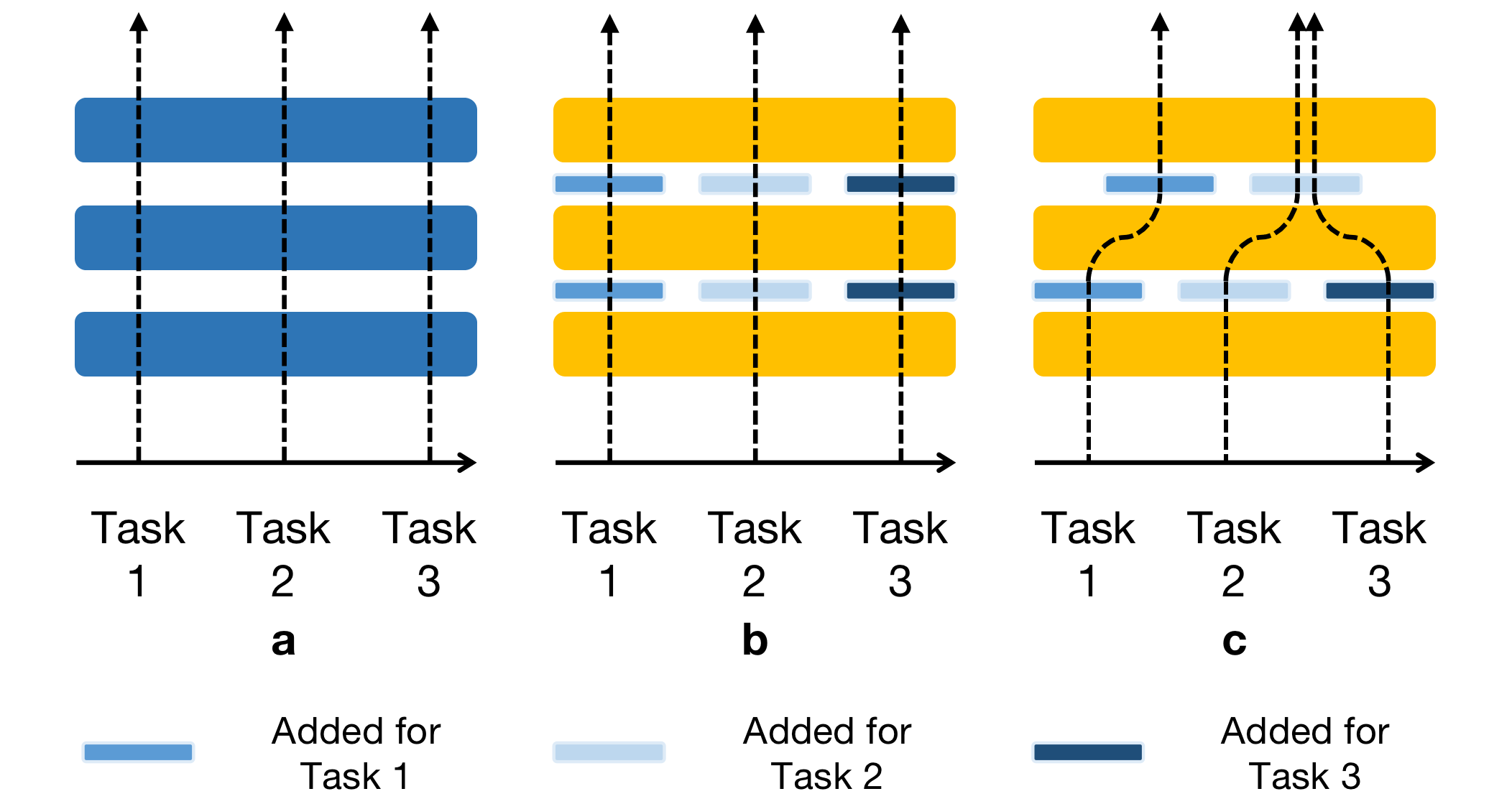}
\end{center}
\caption{Comparison between previous methods (a and b) and our proposed method (c), from a multi-layer transformer model perspective. The blue blocks refer to learnable modules and the yellow blocks refer to frozen pretrained modules . \textbf{a}: retrain the whole model every time when new tasks arrive. \textbf{b}: insert task-specific modules for each task, while keeping the pretrained model frozen. \textbf{c}: detect reusable old modules and add new modules adaptively.
}
\label{fig:compare}
\end{figure}

% An intuitive way to understand catastrophic forgetting is that we optimize old parameters on new data without forcing the model to retain old knowledge, since we only follow the gradients of the loss on current examples. Based on this,
% Based on the method of retaining old knowledge while learning new tasks,
Comparing to continual learning on text classification and question answering \citep{wang2020efficient, holla2020meta, huang2021continual}, continual sequence generation is more challenging, since the output is no longer discrete labels but sequential text data in different styles/domains.
Based on how to retain old knowledge while learning new tasks, current continual sequence generation methods can be categorized into two types. The first one continually learns new tasks on old parameters (Fig \ref{fig:compare} a), with approaches like experience replay \citep{sun2019lamol, chuang2020lifelong} and regularization \citep{mi2020continual} to maintain old knowledge. However, since all tasks share the same parameters, some degree of interference between tasks is unavoidable. 
Another line of work continually inserts new task-specific modules (adapters proposed by \citealp{houlsby2019parameterefficient}) into every transformer layer for every new task while freezing pretrained models and modules used by old tasks (Fig \ref{fig:compare} b, \citealp{madotto2020continual}), which might prevent knowledge transfer between tasks and introduce possible parameter redundancy.
%, since a lot of parameters are isolated and specific to each task instead of shared by similar tasks. 
In this work, we aim to get the best of both worlds: how to encourage the models to reuse modules from previous tasks as much as possible and to only add new modules if needed?

To this end, we propose continual sequence generation with \textbf{\emph{adaptive compositional modules}}, as shown in Fig \ref{fig:compare} c.
% Though we also insert modules into the frozen pretrained model, 
% That is, we add new modules in necessary layers to prevent forgetting \textbf{\emph{after}} detecting reusable old modules to facilitate knowledge transfer.
Specifically, we introduce a two-stage process for every new coming task: a decision stage and a training stage. 
During decision stage, we decide which modules to reuse and whether we need to add a new module. % by \emph{hidden state mixing}. Considering several modules (including the newly added module) in one layer, we pass the weighted average of their outputs to the following network. We finally select the module corresponding to the maximum learned weight. 
During training stage, the model architecture is determined and fixed.  We augment new task's training process with pseudo experience replay \citep{sun2019lamol} to further mitigate forgetting and facilitate knowledge transfer in those shared layers. 
% , 
% This allows our model architecture to func
% This allows us to build our model's architecture 
% in an adaptive and compositional manner.
Our model architecture is \emph{adaptive}, as it 
% (1) \emph{Adaptive}: Our final architecture is adaptive to different types of task sequence: 
% it 
can automatically add new modules for dissimilar tasks and reuse modules for similar tasks, thus making it robust to different scenarios of continual learning. 
Furthermore, it is \emph{compositional} because
for every new task, our new architecture is composed of reused modules from old tasks and newly added modules, which allows  knowledge reuse and transfer. %  at the module level.
% In addition, we use \textbf{hidden state mixing} to decide which modules to reuse and whether we need to add a new module.
% To the best of our knowledge, we are the \emph{first} to propose a dynamic architecture approach for continual sequence generation in NLP.
%

% Concretely, this work takes a closer look at continual learning on sequence generation to evaluate the above adaptive compositional framework. 
% Comparing to continual learning on text classification and question answering \citep{wang2020efficient, holla2020meta, huang2021continual}, continual learning on sequence generation is relatively under-investigated and more challenging, since the output is no longer discrete labels but sequential text data in different styles/domains. 
To evaluate the above adaptive compositional framework,
we experiment with 
% In this work, we decide to take it one step further in this direction by conducting experiments  on 
four representative sequence generation tasks following prior work \citep{sun2019lamol, chuang2020lifelong}:  natural language generation, SQL query generation, summarization and  task-oriented dialogue arriving in a stream. Different from prior work that only tests their methods on very short task sequences or long task sequences with similar tasks only, we validate our approach on longer sequences containing diverse tasks with different levels of similarity.
% all these tasks need to extract useful information from the input sequence and generate the corresponding output sequence, 
% In this case, the model sequentially learn different tasks which vary greatly in their domain and in the pattern of input and output.
We believe this is a suitable scenario to validate both the model's ability to mitigate forgetting and its ability to facilitate knowledge transfer. 
% To sum up, our contributions are two-fold: 
In summary, this work makes two key contributions:
(1) We propose 
% a dynamic architecture approach via 
continual sequence generation with adaptive compositional modules, to maximize knowledge transfer via module-reusing while adaptively adding new modules to mitigate task-interference and catastrophic forgetting.
(2) Experiments with longer and more task sequences show that our approach outperformed  baselines  with higher parameter efficiency.
    % \item We present an analysis showing our compositional framework is able to decide whether to reuse or add modules based on the task-similarity effectively.

\section{Related Work}
% added
\paragraph{Continual Learning} Without allocating new parameters for new tasks, prior work mainly leverages experience replay \citep{wang2019sentence, sun2019lamol} and regularization to mitigate catastrophic forgetting. In experience replay, models are retrained on old examples from previous tasks while learning new tasks. Those old examples are usually stored in a fixed size \citep{mi2020continual} or expanding \citep{huang2021continual} memory buffer. % Known as local adaptation \citep{d2019episodic, wang2020efficient}, experience replay could also be conducted during inference to boost test performance. 
Besides replaying old examples, regularization on the hidden states \citep{wang2019sentence, han2020continual, huang2021continual} or parameters \citep{mi2020continual} could be further added to prevent severe distortion.
Another line of work is to create new parameters for new tasks while freezing parameters used by old tasks. In computer vision, progressive neural network \citep{rusu2016progressive} continually adds new branches of parameters for new image classification tasks with lateral connections to facilitate forward knowledge transfer.  Dynamically expandable network \citep{yoon2017lifelong} expands neural networks at neuron level by using regularization to restrict the number of added neurons. 
While allocating a big network in advance, PackNet \citep{mallya2018packnet} continually assigns a parameter subset to each task by network pruning.% HAT \citep{serra2018overcoming} utilizes task-specific attention mask to assign different parameters for different tasks. 
\citet{li2019learn} employ neural architecture search \citep{liu2018darts} to optimize on new task's structure before learning new tasks. 
In language domain, prior work often utilizes adapter \citep{houlsby2019parameterefficient, madotto2020continual, ermis2022memory}, which could be considered as task-specific MLPs inserted into frozen transformer layers. However, since all adapter modules are designed for only one specific task, no knowledge transfer is directly allowed in this case. Extra modules like attention module \citep{pfeiffer2020adapterfusion}, capsule network \citep{ke2021adapting}, and hypernetworks \citep{jin2021learn} are demonstrated beneficial for knowledge transfer, but they need to introduce extra parameters and fail to consider any reusable or compositional modules. 
% ignore the fact that adapter modules in certain layers are reusable and could be shared by different tasks in continual learning.

Avoiding privacy concerns, this work also follows a line of work that doesn't store real examples for experience replay, such as generating examples by GAN \citep{atkinson2018pseudorecursal}, synthesizing examples \citep{xu2022reality} by model-inversion \citep{smith2021always}, and using unlabeled data in the learning environment \citep{smith2021memory}. In language domain, LAMOL \citep{sun2019lamol} trains the language model to solve current tasks and generate current training examples simultaneously, then this model can generate ``pseudo'' old examples for replay before any new tasks. We adopt this pseudo experience replay along to alleviate the forgetting in the shared modules of our approach.

% \noindent
\paragraph{Continual Learning for Sequence Generation} Building on an auto-regressive language model, LAMOL \citep{sun2019lamol} makes initial exploration on continual sequence generation. On the basis of LAMOL, knowledge distillation \citep{chuang2020lifelong, sun2020distill} is shown to be effective via improving knowledge transfer while changing tasks. ARPER \citep{mi2020continual} combines regularization on parameters \citep{kirkpatrick2017overcoming} with prioritized exemplar replay. Keeping the pretrained model frozen, \citet{madotto2020continual} added task-specific modules for each task together with a perplexity-based classifier, without taking into account the potential for knowledge transfer between different tasks. Instead of blindly adding new modules for new tasks, our approach can detect reusable modules and strategically add new adapter modules in those layers in which reusing old modules would lead to severe forgetting. Without introducing extra knowledge transfer modules, our approach enables knowledge transfer via module sharing.

\paragraph{Task-specific Modules} Traditional finetuning approaches \citep{peters-etal-2018-deep, devlin-etal-2019-bert, radford2019language} usually modify all the parameters in large pretrained modules while learning downstream tasks. Recently, a line of work has been proposed to improve the parameter-efficiency of finetuning by inserting task-specific modules into freezing pretrained models. Adapter \citep{houlsby2019parameterefficient} inserts MLP layers into each transformer layer. PrefixTuning \citep{li2021prefixtuning} prepends key-value pairs to each transformer layer as activations. Prior work also shows that these task-specific modules might benefit from a more adaptive usage. For example, AdapterDrop \citep{ruckle2020adapterdrop} shows that removing adapters from lower transformer layers can almost maintain the original performance while reducing computational overhead. \citet{guo2021adaptive} leveraged latent variables to decide whether to skip adapter modules in certain transformer layers to speed up decoding. However, our approach goes beyond the notion of ``task-specific'', recomposes reusable modules from different tasks, and learns compositional architectures for new coming tasks.

\section{Background}
% \vspace{-0.08in}
\paragraph{Continual Generation Formulation}
% In this work, we focus on continual learning in sequence generation \citep{sun2019lamol, chuang2020lifelong}.  
Assuming multiple sequence generation tasks $\{T_{1}...T_{n}\}$ arrive in a stream, each task $T_{i}$ has a set of training examples $\{P_{1}^{i}, P_{2}^{i} ..., P_{k}^{i}\}$, where $P_{j}^{i}$ denotes a $(input, output)$ pair in Task $i$. While learning on task $T_{i}\ (i > 2)$, we have \textbf{no access} to examples from previous tasks. The final goal is to optimize the model's average performance % \footnote{During Inference, we assume that we know to which task each example belongs to. % , while this is not necessary while \citet{madotto2020continual} show that building an extra classifier could help.
% } 
on \textbf{all} tasks after training on the whole sequence.

\paragraph{Finetuning}
In order to integrate different sequence generation tasks into a single framework, we use \textbf{finetuning} as a general strategy. On the basis of an autoregressive language model, the core idea is to feed the model $input$ and train the model to generate the corresponding $output$ subsequently. To distinguish between tasks, we add an extra $question$ following every $input$ to describe the purpose of each task. For example, the $question$ for natural language generation tasks is \emph{What is the natural language form?}
%and the $question$ for summarization tasks could be \emph{what is the summary ?}. 
Formally, for each $(input, question, output)$ triple, the model is optimized to generate the corresponding $output$ given $input$ and $question$:
\begin{equation*}
    \begin{split}
        L_{finetune}(x) = \sum_{t=m+1}^{n} - \log P(x_{t} | x_{<t})
    \end{split}
\end{equation*}
where $x=\{x_{1},...,x_{n}\}$ denotes the concatenation of $input$, $question$ and $output$, and $\{x_{1},...,x_{m}\}$ refers to $input$ and $question$.

\paragraph{Adapter} The \emph{module} used in our framework refers to adapter \citep{houlsby2019parameterefficient}, which is a task-specific module inserted into each frozen pretrained transformer layers \citep{NIPS2017_3f5ee243}. 
%Training adapter only is demonstrated effective enough to almost maintain the performance comparing to finetuning the whole model. 
In addition to residual connection \citep{7780459} and layer normalization \citep{ba2016layer}, one transformer layer contains two primary sub-layers: an attention layer and a feed forward layer. One adapter module consists of two multi-layer perceptrons ($MLP$), one ($MLP_{MH}$) following the multi-head attention layer and one ($MLP_{FF}$) following the feed forward layer.
%Every inserted MLP follows a bottle-neck architecture: it first projects features into a lower dimension, then project back to the original dimension. 

\section{Two-Stage Methods}
Motivated by prior continual sequence generation work \citep{madotto2020continual} that uses Adapter \citep{houlsby2019parameterefficient} to insert new adapter module into every transformer layer for each new coming task, 
we propose to strategically 
\emph{decide} whether we can reuse some adapter modules from old tasks before \emph{training} on each new coming task, in a two-stage manner:  decision stage and training stage, where the former determines the architecture for new tasks and the later trains the model. % on new tasks.

\begin{figure*}[t]
\begin{center}
\includegraphics[width=1.0\linewidth]{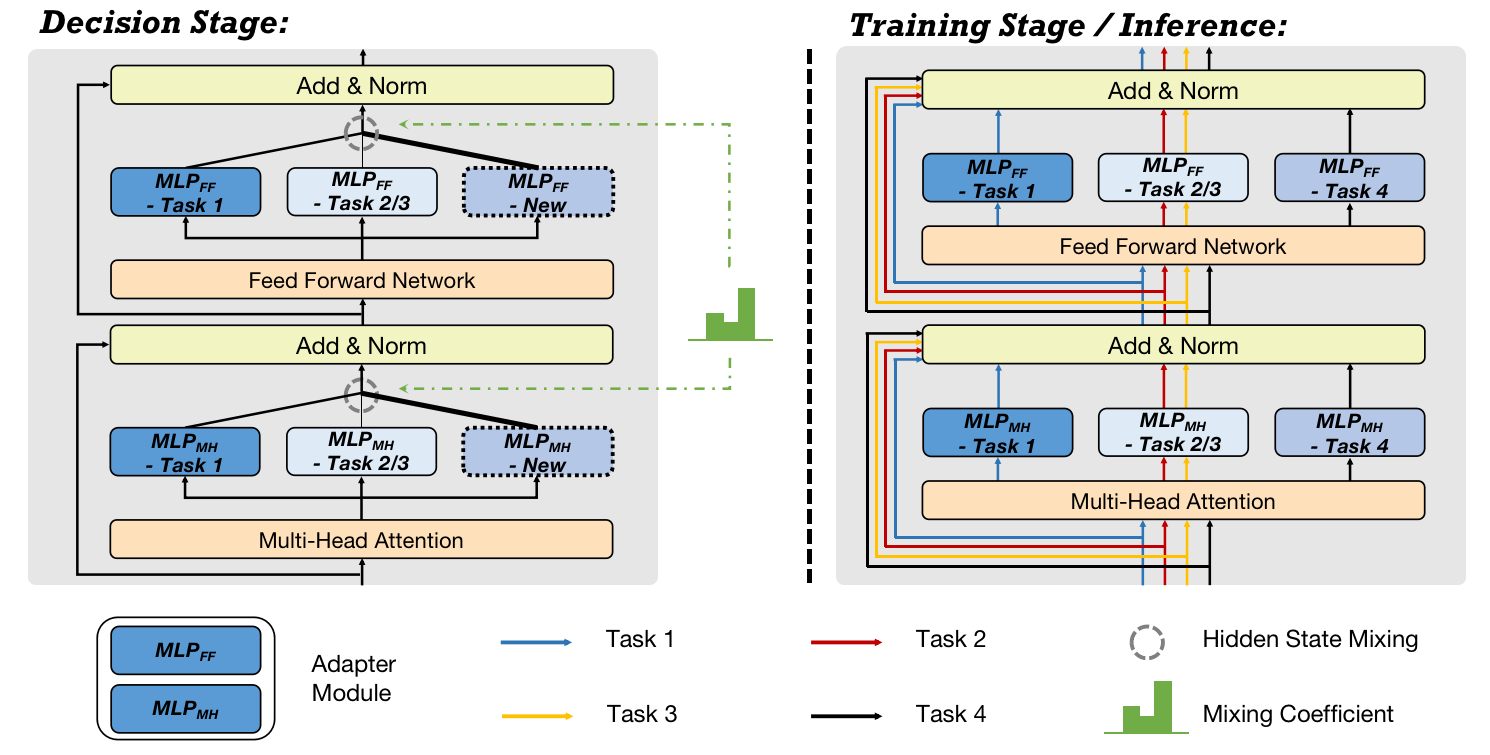}
\end{center}
\caption{Our proposed model architecture with adaptive compositional modules for transformer layers. Assume after learning three tasks (1, 2, 3), we have one module for task 1, and another for task 2 and 3 in this layer. \textbf{Left}: During decision stage for task 4, we first insert a new module at this position, then all inserted modules will be used for selection using hidden state mixing. \textbf{Right}: Assume that we finally decide to add one module at this position, then each task would use its own architecture during training stage and inference. 
}
\label{fig:model}
\end{figure*}

\subsection{Decision Stage}
The decision stage aims to answer two questions: 
% In decision stage, for each layer, we try to solve two problems: 
do we need to add a new module in this layer? If not, which old modules should we reuse? Inspired by interpolation-based data augmentation \citep{chen2020mixtext, chen2021empirical} and neural architecture search \citep{liu2018darts}, we utilize \textbf{Hidden State Mixing} for module selection. 
Assume that there are several modules as potential candidates to be selected, 
%when passing the output from lower layers to these perceptrons,
after calculating their output separately,
we calculate their \emph{weighted average} as the overall output, which is then passed to the next part of the model (See the left part in Figure \ref{fig:model}). After training the entire model end-to-end, we assume that the module with the largest learned weight is the most useful one, and thus will be selected for the reuse. 

% Here the weight coefficient is learnable. 
Formally, assume that we already have inserted $k$ modules into the $l$th transformer layer, each consisting of two MLPs: $(MLP_{MH}^{1, l}, MLP_{FF}^{1, l})$... $(MLP_{MH}^{k, l}, MLP_{FF}^{k, l})$. At the beginning of decision stage, we add one more module $(MLP_{MH}^{k+1, l}, MLP_{FF}^{k+1, l})$. Given these learnable weight coefficients $[\lambda_{1, l}, \ldots,  \lambda_{k+1, l}]$, multi-head attention layer output $o_{mh}^{l}$, the feed forward layer output $o_{ff}^{l}$, we mix the hidden states as follow:
\begin{equation*}
\begin{split}
\begin{aligned}
        h_{mh}^{l} & = \sum_{t=1}^{k+1} \lambda_{t, l}MLP_{MH}^{t, l}(o_{mh}^{l}) \\
        h_{ff}^{l} & = \sum_{t=1}^{k+1} \lambda_{t, l}MLP_{FF}^{t, l}(o_{ff}^{l})
\end{aligned}
\end{split}
\end{equation*}
where both $h_{mh}^{l}$ and $h_{ff}^{l}$ are then fed into their following \textsl{Add \& Norm} layers.
%$h_{mh}^{l}$ is the input for the Add \& Norm layer after multi-head attention layer and $h_{ff}^{l}$ is the input for the Add \& Norm layer after feed forward layer.
To ensure $\sum_{t=1}^{k+1} \lambda_{t, l} = 1$, we use \textsl{softmax} function to produce $\lambda_{1, l}, \ldots, \lambda_{k+1, l}$ from $c_{1, l}, \ldots, c_{k+1, l}$:
\begin{equation*}
    \begin{split}
        \lambda_{i, l} = \frac{e^{c_{i, l}}}{\sum_{t=1}^{k+1} e^{c_{t, l}}}, i = 1 \dots k+1
    \end{split}
\end{equation*}

Using this mixing approach in every transformer layer,  we optimize our model using $L_{train}$ (see Sec \ref{training stage}) for the new task and find the most suitable modules for each layer. Note that (i) In this process, the pretrained model and all old modules are frozen, and only mixing coefficients and newly added modules will be learned. (ii) Calculating the weighted average is a convenient approximation of using one adapter at a time, which is the real setting during training stage and inference. (iii) Comparing to other baselines in Figure \ref{fig:compare}, introduced decision stage to decide the architecture does introduce extra computation, while computation of different MLPs at one position is parallelizable to speed up.   % (2) Since we treat the two multi-layer perceptron as a whole, we only need to assign one group of weight coefficient for each layer.

%\begin{comment}
% \yanzhe{Still doing ablation study on Entropy loss to study its contribution}
% \paragraph{Entropy Loss}
To avoid the learned weight coefficient $\lambda_{1, l}, \ldots, \lambda_{k+1, l}$ to be too close to a uniform distribution in certain layers, we further add
an additional regularization term to $L_{train}$, which is the sum of \textbf{\emph{entropy}} of every discrete probability distribution $[\lambda_{1, l}, \dots,\lambda_{k+1, l}]$: % , and then  and minimize the entropy of $[\lambda_{1}, \dots,\lambda_{k+1}]$ to encourage a non-uniform weight distribution:
\begin{equation*}
    \begin{split}
        L_{entropy} = \gamma \sum_{l} \sum_{i=1}^{k+1} - \lambda_{i, l} \log (\lambda_{i, l})
    \end{split}
\end{equation*}
where $\gamma$ is a coefficient tuned as a hyper-parameter.
% Taken it together, the final training loss is $L = L_{train} + \gamma \sum L_{entropy}$, where we sum up the entropy loss for each layer and $\gamma$ is a hyper-parameter. We will introduce $L_{train}$ in sec \ref{training stage}.
%\end{comment}
% \paragraph{Weight Initialization}

In this stage, a trivial solution could be allocating a new module in every layer regardless of whether old modules are reusable. % , which will result in allocating different modules for different tasks. 
To avoid this trivial solution and reuse shareable modules as much as possible, we design a prior using the \textbf{\emph{ initialization of the coefficient weights}}. For every $l$, $c_{1, l}...c_{k, l}$ is initialized to $c\ (c>0)$, while $c_{k+1, l}$ is initialized to $-c$. After \textsl{softmax}, the weight of each old module is $e^{2c}$ times the weight of the new module, increasing the tendency to reuse old modules.
% \vspace{-0.08in}
\subsection{Training Stage}
\label{training stage}
% Via reusable models, our approach enables module sharing. 
% Since our approach permits module sharing, 
We further incorporate pseudo experience replay \citep{sun2019lamol} to mitigate forgetting and facilitate knowledge transfer in those shared modules. The main idea is to teach a generative model to solve current task and to generate current task's examples \textbf{simultaneously}. Then before training on each new task, we can generate a set of pseudo old examples and replay them during training. 

Thus, in addition to the finetuning loss to solve each task, we introduce an extra loss $L_{gen}$ for the model to generate current task's examples. Formally, given the whole sequence of $x=\{input, question, output\}$, we first add a special token \texttt{[GEN]} at the beginning of $x$ to form a new sequence $x^{\prime}$, and then optimize the model as follows:
\begin{equation*}
    \begin{split}
        L_{gen}(x^{\prime}) = \sum_{t=1}^{n+1} - \log P(x^{\prime}_{t} | x^{\prime}_{<t})
    \end{split}
\end{equation*}
Note that we use different special tokens for different tasks, thus we can generate examples for specified tasks afterwards. Combining with the finetune loss, the overall training loss is:  
\begin{equation*}
    \begin{split}
        L_{train} = L_{finetune} + \eta L_{gen}
    \end{split}
\end{equation*}
where $\eta$ is the weight for the $L_{gen}$ loss.

% We have different training strategies for different decisions made from the decision stage. (i) When no old module is reused, i.e., all modules used by the new task are newly added. In this case no learnable module is shared between the old tasks and the current one, there is less likely a problem of forgetting and  less knowledge transfer. In this case, there is no need for pseudo experience replay so we only train our model using $L_{train}$ on the current dataset. (ii) When some old modules for the new task are reused, i.e., certain modules are shared between old tasks and the new task. Since our model is trained on $L_{gen}$ in previous tasks, we first generate pseudo training examples of those tasks which at least share one module with the current task, and then train our model using $L_{train}$ on the current dataset together with the generated examples.
Once our model has the ability to generate ``pseudo`` examples from old tasks, another question is \emph{When to generate ``pseudo`` examples?} Since those ``pseudo`` examples are for shared modules between old tasks and the current task, we only generate them while some old modules are reused for the current task. In that case, we train our model using $L_{train}$ on the current dataset together with the generated examples. Otherwise, there is no need for pseudo experience replay and we just train our model using $L_{train}$ on the current dataset.

\section{Experiments}
% \vspace{-0.08in}
\subsection{Datasets}
Following \citet{sun2019lamol} and \citet{chuang2020lifelong}, we evaluate our approach on four representative sequence generation tasks: natural language generation, SQL query generation, summarization and task-oriented dialogue modeling. 
%\footnote{Datasets available at: \\ \url{https://github.com/chho33/LAMOL} \\ \url{https://github.com/voidism/L2KD}}.
% Considering the application of continual learning in real world scenarios, 
Specifically, we test our proposed approach under two common scenarios: (1) \emph{CL on similar tasks}: in this case, the new coming tasks often share the same task pattern with learned tasks, but are from different domains. % Beyond mitigating catastrophic forgetting, continual learning methods also need to detect and enable potential knowledge transfer under this situation. 
We use E2ENLG \citep{novikova2017e2e} and four different domains (restaurant, hotel, tv, laptop) from RNNLG \citep{wen-etal-2015-semantically} to form five \emph{similar} tasks. % Since all these five tasks are structured text to natural language generation, sharing a similar task pattern but with different domains/intents. 
Then we use four different orders of these tasks as our testing task sequences. 
(2) \emph{CL on dissimilar tasks}: in this case, the distribution shift between new tasks and old tasks could be relatively large, so the major challenge is to retain old knowledge as much as possible while learning new tasks. In this case, we further incorporate WikiSQL (SQL query generation, \citealp{zhong2017seq2sql}), CNN/DailyMail (news article summarization \citealp{see2017get}), MultiWOZ (semantic state sequence generation \citep{budzianowski2018multiwoz}) into our task sequences\footnote{We use ``e2e'' for E2ENLG, ``rest'' for RNNLG (restaurant), ``hotel'' for RNNLG (hotel), ``tv'' for RNNLG (tv), ``laptop'' for RNNLG (laptop), ``wiki'' for WikiSQL, ``cnn'' for CNN/DailyMail, ``woz'' for MultiWOZ. }. % In this case, the task sequences not only include similar natural language generation tasks, but also include dissimilar tasks with diverse pattern of input and output. 
% In this case, our task sequences contain both similar tasks and dissimilar tasks.
We randomly pick four different orders as our testing task sequences. In total, we use eight different task sequences (Table \ref{table-order}) to evaluate our models. The statistics/metrics for each dataset and the finetuing results are in Appendix \ref{sec:appendix}.

\begin{table}
\centering
\begin{tabular}{cl}
\toprule
\textbf{Order} &  \textbf{Task Sequence} \\
\midrule
1 & e2e $\shortto$ rest $\shortto$ hotel $\shortto$ tv $\shortto$ laptop \\
2 & laptop $\shortto$ tv $\shortto$ hotel $\shortto$ rest $\shortto$ e2e \\
3 & rest $\shortto$ tv $\shortto$ e2e $\shortto$ laptop $\shortto$ hotel \\
4 & hotel $\shortto$ e2e $\shortto$ rest $\shortto$ laptop $\shortto$ tv \\
5 & woz $\shortto$ cnn $\shortto$ e2e $\shortto$ rest $\shortto$ hotel \\
6 & e2e $\shortto$ wiki $\shortto$ hotel $\shortto$ woz $\shortto$ rest \\
7 & hotel $\shortto$ e2e $\shortto$ woz $\shortto$ wiki $\shortto$ cnn \\
8 & cnn $\shortto$ hotel $\shortto$ wiki $\shortto$ e2e $\shortto$ woz  \\
  
\bottomrule
\end{tabular}
\caption{\label{table-order}
Eight random different task sequences. The first 4 includes different orders of similar tasks, the last 4 includes different orders including dissimilar tasks.
}
\end{table}
% \vspace{-0.1in}

\begin{table*}[t]
\centering
\small
\begin{tabular}{ll| p{1.3cm}<{\centering\arraybackslash} p{1.3cm}<{\centering\arraybackslash} p{1.3cm}<{\centering\arraybackslash}| p{1.3cm}<{\centering\arraybackslash} p{1.3cm} <{\centering\arraybackslash} p{1.3cm} <{\centering\arraybackslash} p{1.3cm}<{\centering\arraybackslash}}
\toprule
\multicolumn{2}{l|}{Methods} &
  \textbf{Finetune} &
  \textbf{EWC} &
  \textbf{LAMOL} &
  \textbf{\begin{tabular}[c]{@{}l@{}}Adapter\\ +CL\end{tabular}} &
  \textbf{\begin{tabular}[c]{@{}l@{}}Adapter\\ +Drop\end{tabular}} &
  \textbf{\begin{tabular}[c]{@{}l@{}}Adapter\\ +LAMOL\end{tabular}} &
  \textbf{Ours}  \\ \midrule
\multicolumn{2}{l|}{\begin{tabular}[c]{@{}l@{}}Pseudo \\ Experience Replay\end{tabular}} & \xmark & \xmark & \cmark & \xmark & \xmark & \cmark & \cmark \\ \midrule
\multirow{4}{*}{Similar Tasks}                            & \# 1                        &43.0 & 56.9 & 66.3  &  64.2&   63.9 &  65.9&  \textbf{66.1}  \\
                                                          & \# 2                        &37.0 & 47.9& 67.0  &  64.2&   63.9&  66.2&  \textbf{66.5}  \\
                                                          & \# 3                        &51.7  & 61.4& 66.6 &  64.2&   63.9&  65.6& \textbf{65.8}  \\
                                                          & \# 4                        &45.0  & 58.3 & 66.6 &  64.2&   63.9&  65.2&  \textbf{65.7}   \\ \hline
\multicolumn{2}{l|}{Avg Performance}                                                             & 44.2 & 56.2 & 66.6 &  64.2 &  63.9 & 65.7 & \textbf{66.0}  \\
\multicolumn{2}{l|}{Avg Learnable Para.}                                                             &  124.45M & 124.45M & 124.45M &  8.95M & 6.71M & 1.79M & 2.44M  \\
\midrule
\multirow{4}{*}{Dissimilar Tasks}                         & \# 5                        &33.6 & 37.5&  57.0  &  57.5&   57.4&  54.3& \textbf{58.2} \\
                                                          & \# 6                        &32.6 & 37.9& 62.5  &  64.9&   64.5&  62.2&  \textbf{65.9}  \\
                                                          & \# 7                        &19.7 & 37.5 & 56.7  &  57.3&   56.7&  54.6&  \textbf{58.3}  \\
                                                          & \# 8                        &26.3 & 38.8 & 56.8 &  57.3&   56.7&  53.8&  \textbf{58.2}  \\ \hline
\multicolumn{2}{l|}{Avg Performance}                                                             & 28.1 & 37.9 & 58.3 & 59.3 &  58.8 & 56.2 & \textbf{60.1}  \\
\multicolumn{2}{l|}{Avg Learnable Para.}                                                             & 124.45M & 124.45M & 124.45M &  8.95M & 6.71M & 1.79M &  6.60M  \\ \bottomrule
\end{tabular}
\caption{\label{table-res1}
% Summary of final performance. For each method on each task sequence, we report 
The mean of final performance score on all tasks. We use two random seeds for each task sequence. Note that the final performance of Adapter+CL and Adapter+Drop is not affected by task ordering within the same group of tasks. For each sequence, we mark the best representation in \textbf{bold}, where LAMOL is not compared due to the difference in the order of magnitude of the learnable parameters. For each scenario, the $p$-values of paired $t$-test between 8 numbers of our approach and the second highest comparable baseline is smaller than 0.05, demonstrating significant improvement.
}
\end{table*}

\subsection{Baselines}
We compare our proposed model with the following baselines: (i) \textbf{Finetune} \citep{yogatama2019learning}: We finetuned GPT-2 model on several tasks sequentially.
(ii) \textbf{EWC} \citep{kirkpatrick2017overcoming} added regularization on parameters according to their importance to old tasks.
(iii) \textbf{LAMOL} \citep{sun2019lamol} finetuned the whole GPT-2 model continually with the help of pseudo experience replay.
(iv) \textbf{Adapter+CL} \citep{madotto2020continual} inserted adapter \citep{houlsby2019parameterefficient} modules into every GPT-2's layer for each task.
(v) \textbf{Adapter+Drop} \citep{ruckle2020adapterdrop}: We removed all those adapter modules from the first three layers in GPT-2 based on Adapter+CL.
(vi) \textbf{Adapter+LAMOL}: We only inserted adapter modules into every transformer layer for the first task, then used those adapter modules to learn the whole the task sequence with pseudo experience replay. 
Note that ARPER \citep{mi2020continual} also tackles the problem of continual sequence generation, but it needs an extra memory buffer to store examples from old tasks, which is not comparable with ours.
% \vspace{-0.02in}

\paragraph{Implementation Details}
We use GPT-2 \citep{radford2019language} in HugginceFace Transformers \citep{wolf2020transformers} as our backbone and adapter implementation by AdapterHub \citep{pfeiffer-etal-2020-adapterhub}.
% All experiments are conducted on NVIDIA RTX 2080 Ti with 11GB memory with a maximum batch size of 4. We use AdamW \citep{loshchilov2017decoupled} as our optimizer. We set the learning rate $lr=1.75e-4$ for all tasks except WikiSQL, and $lr=3e-4$ for WikiSQL. For epochs, we select the best setting of each task from $\{9, 12, 15\}$. Weight initialization parameter $c$ is selected from $\{0.03, 0.05, 0.07, 0.10, 0.12, 0.15\}$, loss coefficient $\gamma$ is set to $0.01$, $\eta$ is set to $0.25$. % For sequence decoding, we follow the setting used in \citet{sun2019lamol} and 
More details can be found  in Appendix \ref{sec:appendix}.

%\vspace{-0.08in}
\section{Results and Analysis}
%\vspace{-0.05in}
\begin{figure*}[ht]
\begin{center}
\includegraphics[width=1.0\linewidth]{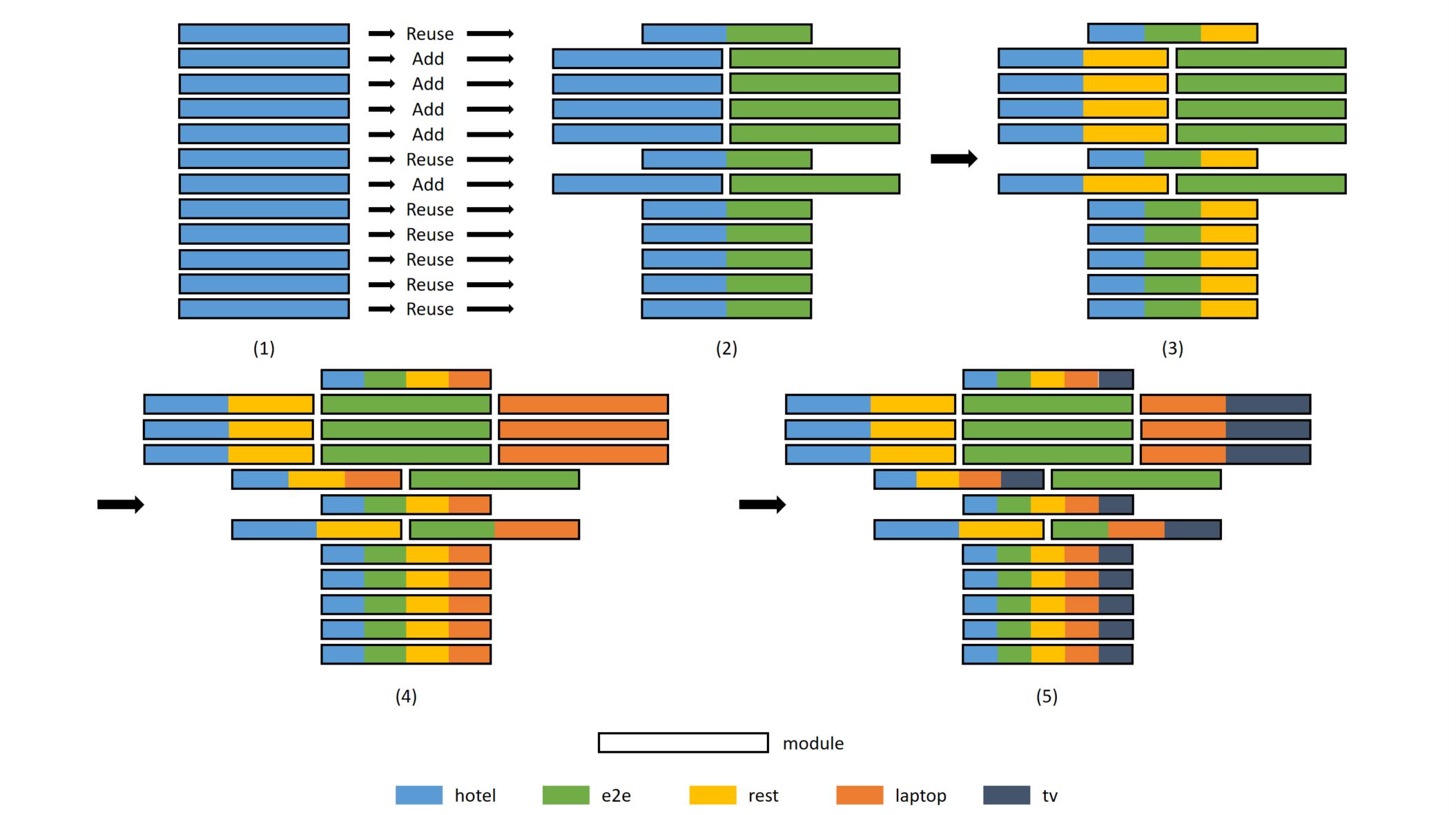}
\end{center}
\caption{The growing process of our model on sequence: hotel $\shortto$ e2e $\shortto$ rest $\shortto$ laptop $\shortto$ tv. The 1st layer is shown at the bottom and the 12th layer is at the top of each figure. Note that here we only depict the architecture growing process of our inserted modules: (i) Each rectangle represents a module \textit{added} in that specific transformer layer.  (ii) Each module is painted with the corresponding color if it is used by a task. (iii) Modules with multiple colors are \textit{shared} by multiple tasks.
}
\label{fig:arch-growth}
\end{figure*}

% Given a task sequence, 
To evaluate the overall performance on all tasks, we use the mean of all tasks' performance score following \citet{sun2019lamol, mi2020continual, madotto2020continual}.  For each scenario (\emph{similar} tasks and \emph{dissimilar} tasks), we report the average of mean scores on all sequences as an overall metric. Beyond these, we also provide (i) evaluation results using \emph{geometric mean} and (ii) final performance of each task in Appendix \ref{sec:appendix}. % To be more comprehensive, we also provide full results in Appendix to further support our discussion.
Table \ref{table-res1} summarizes the final performance on all eight task sequences. We observed that finetuning sequentially suffered from very severe forgetting, no matter on \emph{similar} or \emph{dissimilar} tasks, highlighting the importance of continual learning work. Though EWC can significantly increase the performance of finetuning, its performance is still far behind LAMOL, highlighting the importance of experience replay.

\begin{table}[t]
\centering
\small
\setlength\tabcolsep{3pt}

\begin{tabular}{l|cc|cc}
\toprule
\multirow{2}{*}{\textbf{Method}} & \multicolumn{2}{c}{Sequence \#1}                    & \multicolumn{2}{|c}{Sequence \#8}                    \\
                                 & Avg                  & \multicolumn{1}{c|}{Avg L.P.} & Avg                  & \multicolumn{1}{c}{Avg L.P.} \\
\midrule

Ours & 66.1 & 2.24M & 58.2 & 6.49M\\
w/o Entropy loss & 66.1 & 2.54M & 57.6 & 6.49M\\
w/o Weight Ini & 64.2 & 7.09M & 57.7 & 8.65M\\
w/o Pseudo ER &  43.2 & 2.08M & 55.9 & 6.34M\\

\bottomrule
\end{tabular}
\caption{Ablation study on (i) entropy loss (ii) weight initialization (iii) pseudo experience replay. The left part includes results for sequence \#1 while the right part includes result for sequence \#8. Note that ``Avg`` refers to the mean of performance score on all tasks and ``Avg L.P.`` refers to the mean of learnable parameters.}
\label{table:ablation-brief}
\end{table}

\begin{table}[ht]
\centering
\small
\begin{tabular}{lccc}
\toprule
\textbf{Length} & \multicolumn{1}{l}{\textbf{\begin{tabular}[c]{@{}l@{}}Adapter\\ +CL\end{tabular}}} & \multicolumn{1}{l}{\textbf{\begin{tabular}[c]{@{}l@{}}Adapter\\ +LAMOL\end{tabular}}} & \textbf{Ours} \\ \midrule
2 Tasks(\#1) &  56.8 (+0.0)&  57.5 (+0.8) &  57.7 (+0.9)\\
3 Tasks(\#1) &  59.5 (+0.0)&  60.3 (+0.6) &  60.1 (+0.5)\\
4 Tasks(\#1) &  62.3 (+0.0)&  63.5 (+1.3) &  63.7 (+1.6)\\
5 Tasks(\#1) &  64.2 (+0.0)&  65.9 (+2.0) &  66.1 (+2.1)\\ \hline
2 Tasks(\#8) &  45.4 (+0.0)&  46.2 (+1.3) &  46.0 (+1.2) \\
3 Tasks(\#8) &  51.3 (+0.0)&  51.9 (+0.8) &  52.3 (+0.9) \\
4 Tasks(\#8) &  50.9 (+0.0)&  49.7 (-1.7) &  51.8 (+0.6) \\
5 Tasks(\#8) &  57.3 (+0.0)&  53.8 (-4.6) &  58.2 (+0.5) \\ \bottomrule
\end{tabular}
\caption{\label{table-length}
Impact of the task sequence length. Note that ``n Tasks(\#i)'' refers to after sequentially training on the first n tasks in sequence \#i, we report the mean of performance score on those n tasks and the backward transfer in parentheses.
}
\end{table}

For sequences containing \emph{similar} tasks, the performance of Adapter+CL is inferior to Adapter+LAMOL even with more learnable parameters. This indicates that %parameter isolation is not always the solution: 
sharing parameters and experience replay can further facilitate knowledge transfer when tasks are similar. 
On the premise of pseudo experience replay, our method performs better than Adapter+LAMOL, demonstrating the effectiveness of our adaptive compositional architecture. Our approach also achieves a much higher parameter efficiency than Adapter+CL and Adapter+Drop. 
For sequences containing \emph{dissimilar} tasks where the transferable knowledge is limited and parameter sharing might cause degradation,  Adapter+CL and Adapter+Drop seem more robust  compared to Adapter+LAMOL and LAMOL, since they avoid catastrophic forgetting by parameter isolation. Using a similar number of parameters to Adapter+Drop, our method outperforms Adapter+CL consistently on all task sequences, confirming that our method can prevent interference between dissimilar tasks while 
reducing parameter redundancy.
% Overall, our approach achieves the most robust results in both two scenarios, while achieving higher parameter-efficiency than LAMOL and Adapter Baselines.

\begin{table*}[t]
\centering
\begin{tabular}{ll}
\toprule
\multicolumn{2}{l}{\textbf{E2E NLG (\#1):} name{[}Strada{]}, eatType{[}coffee shop{]}, area{[}city centre{]}}                                                                                                              \\ \midrule
\multicolumn{1}{l}{Reference}     & \textit{There is a coffee shop in the city centre called the Strada.}                                                                                                               \\ \hline
\multicolumn{1}{l}{Adapter+CL}    & \textit{Strada serves coffee, \red{is a nice coffee shop}, in city centre.}                                                                                                          \\ \hline
\multicolumn{1}{l}{Adapter+LAMOL} & \textit{Strada is a coffee shop \orange{serving city centre food}}             \\ \hline
\multicolumn{1}{l}{Ours}          & \textit{Strada is a coffee shop located in the city centre.}                                                                                                           \\ \toprule
\multicolumn{2}{l}{\textbf{WikiSQL (\#8):} which team has pick 13 in round 2 ?}                                                                                                                          \\ \midrule
\multicolumn{1}{l}{Reference}     & \textit{select team from table where round = 2 and pick = 13}                                                                                                                             \\ \hline
\multicolumn{1}{l}{Adapter+CL}    & \textit{select team from table where pick = 13 and round = \red{round} 2}                                                                                                                                                                     \\ \hline
\multicolumn{1}{l}{Adapter+LAMOL} & \textit{select team from table where round = 2 \blue{(missing: and pick = 13)}} \\ \hline
\multicolumn{1}{l}{Ours}          & \textit{select team from table where pick = 13 and round = 2}                                                                                                                             \\ \bottomrule
\end{tabular}
\caption{Output comparison after training on sequence \#1 and \#8. We visualized e2e and wiki as two representative tasks and color redundant information in \red{red}, missing information in \blue{blue} and grammar mistakes in \orange{orange}.}%\vspace{-0.1in}
\label{table:case study}
\end{table*}
% \vspace{-0.08in}

% \vspace{-0.1in}
\subsection{Ablation Studies}% \vspace{-0.08in}
% We evaluated the contribution of each components in our approach by
% Our ablation studies used one random task sequence from each scenario: sequence \#1  and \#8.
We randomly selected task sequence \#1 from \emph{similar} tasks and sequence \#8 from sequences of \emph{dissimilar} tasks for our ablation studies. 

\paragraph{Importance of Each Component}
To examine the importance of each component in our method, we experiment with different settings: 
not using entropy loss (w/o Entropy Loss), initializing all weight coefficients with zero (w/o Weight Ini), and not replaying pseudo data (w/o Pseudo ER). % To have a comprehensive understanding. 
As shown in Table \ref{table:ablation-brief}, 
% (detailed results in Appendix \ref{sec:appendix} Table \ref{table:ablation}), 
we found that (i) After removing entropy loss, the performance on sequence \#1 is almost maintained by using more parameters. Meanwhile, the performance on sequence \#8 drops significantly while using the same number of parameters. This observation suggests that the entropy loss is beneficial to achieve a better trade-off between adding parameters and maintaining good performance. 
% could be attributed to the fact that more modules are allocated for sequence \#8 than sequence \#1, thus the entropy loss is more effective in selecting reusable modules.
(ii) When we initialize all weight coefficients with zero, there is no explicit tendency to reuse old examples. In this case, many redundant modules are created thus preventing knowledge transfer, which leads to performance drop on both sequences. 
The drop on sequence \#1 is more severe due to there is more transferable knowledge between similar tasks. We therefore conclude that weight initialization is important to enable knowledge transfer between similar tasks.
(iii) Removing pseudo experience replay leads to the most severe performance drop on both sequences. Though our approach strategically detect which modules can be reused, directly training them on new tasks without protecting old knowledge will lead to catastrophic forgetting.

\paragraph{Impact of Task Sequence Length}
Prior work in continual learning \citep{madotto2020continual, huang2021continual} suggests that the differences in sequence length could influence the performance of continual learning. %To gain better understanding of the learning process and make a thorough comparison between various methods, 
To this end, we further investigated the impact of sequence length in Table \ref{table-length}, where we reported the average performance at every step and calculated \emph{Backward Transfer} following \citet{lopez2017gradient}:
\begin{equation*}
    \begin{split}
        & BWT_{k} = \frac{1}{k - 1}\mathbb{E}_{i = 1 \dots k - 1} (R_{k,i} - R_{i,i})
    \end{split}
\end{equation*}
where $R_{i,j}$ is the performance score on the $j$th task
after training on the $i$th task.

We found that, on sequence \#1, Adapter+LAMOL and our method consistently outperform Adapter+CL in all stages, which could be explained by better knowledge transfer between multiple tasks. Beyond that, our method outperforms Adapter+LAMOL in most cases, demonstrating the benefits of adaptively adding modules.
On sequence \#8, %while 
Adapter+LAMOL % is still competitive after learning two or three tasks, it
struggles when the length of task sequence becomes longer. As more and more tasks arrive, the impact of task dissimilarity and distribution shift gets larger that pseudo experience replay cannot cope with. In that case, there is limited backward transfer but severe forgetting. In contrast, Adapter+CL and our method demonstrate their robustness after learning more tasks in a stream. Our method also outperforms Adapter throughout the learning process, demonstrating we can enable knowledge transfer even the similarity between tasks is limited.
%\vspace{-0.08in}

\paragraph{Case Study}
We selected e2e in sequence \#1 and wiki in sequence \#8 as two representative tasks to illustrate the final output generated by different approaches in Table \ref{table:case study}. After training on the whole sequence, Adapter+LAMOL cannot correctly convey the information provided in the input, suffering from generating grammar mistakes and missing key points. This could be attributed to the interference from learning new coming tasks. While Adapter+CL successfully mitigate this problem by parameter isolation, our approach works similarly using less parameters and generates better sequences without redundant information.

\subsection{The Growth of Compositional Modules} %\vspace{-0.08in}
% To facilitate the analysis, we use the cosine similarity between each task's word frequency distribution to measure task similarity, which is reported in Fig \ref{fig:task-sim} in Appendix \ref{sec:appendix}. 
To illustrate the process of adding/reusing modules, we depict the model architecture at each stage in Fig \ref{fig:arch-growth} using sequence \#4, which is the most challenging sequence containing \emph{similar} tasks according to Table \ref{table-res1}.
% , on which the model learn different NLG tasks in the order of: hotel $\shortto$ e2e $\shortto$ rest $\shortto$ laptop $\shortto$ tv. 
%On this sequence, according to our design intention, our approach need to achieve the following simultaneously: (i) create necessary new modules when the distribution shift between tasks is relatively big (e.g.,  hotel $\shortto$ e2e). (ii) find modules used by similar previous task and reuse those that can be shared (e.g., e2e $\shortto$ rest). (iii) add as few modules as possible when the distribution shift betweek tasks is relatively small (e.g., laptop $\shortto$ tv). 
% We observed that, 
%after learning the first task (hotel), we have 12 adapter modules inserted into each transformer layer. 
Since the similarity between the second task (e2e) and the first task (hotel) is low (see Figure \ref{fig:task-sim} in Appendix \ref{sec:appendix}), our framework automatically learns to add extra adapter modules in layer $\{6,8,9,10,11\}$ before training on the second task. When the third task (rest) arrives, given its high similarity to the first task, our method correctly decides to reuse all modules used in the first task. Interestingly, the architecture for the fourth task is composed of shared modules with the first 3 tasks in layer $\{1, 2, 3, 4, 5, 7, 12\}$, shared module with the second task in layer $6$, shared the module with the first and the third task in layer $8$, and also added new modules for the fourth task in layer $\{9, 10, 11\}$. For the fifth task, our method reuses all modules used by the fourth tasks due to their high similarity. This demonstrates that our method is adaptive to different incoming tasks and is able to compose modules from different old tasks for new tasks. We also provide a comparison in Appendix \ref{appendix: module comparison} to demonstrate the effect of reusing modules from different transformer layers.

\section{Conclusion}

This work examined continual sequence generation with adaptive compositional modules, where we proposed hidden state mixing to adaptively compose old and new modules for new tasks and utilized pseudo experience replay to facilitate knowledge transfer. 
Experiments conducted on various sequence generation tasks demonstrated that our method achieves better performances with higher parameter efficiency over previous state-of-the-art baselines, both on \emph{similar} task sequences and \emph{dissimilar} task sequences. 
Our work is also subject to a few limitations such as the introduced extra training time. In the future, we plan to investigate how to further speed up the decision stage more efficiently and generalize 
the current framework to more diverse NLP tasks such as text classification and machine  translation. 
% to task sequences containing more diverse NLP tasks.

% In the future, % (i) we plan to make the task sequence more diverse by adding text classification and question answering. That will put higher demands on the adaptability of the method. 
% instead of randomly selecting several task sequences for evaluation, we will highlight the importance of \emph{task sequence design}. By carefully designing representative task sequences, we can increase the evaluation efficiency and produce more convincing comparison.  Besides module weights, we believe that gradient information \citep{kirkpatrick2017overcoming, lopez2017gradient} could be further considered in the decision stage. Considering that gradient information is shown helpful in studying knowledge transfer and interference \citep{riemer2018learning}, this might be beneficial to improve robustness and interpretability of our Adaptive Compositional Modules.

\section*{Acknowledgment}
We would like to thank the anonymous reviewers
for their helpful comments, and the members of Georgia Tech SALT group for their feedback. This work is funded in part by Salesforce and Cisco.

\bibliography{anthology,custom,tacl2018}
\bibliographystyle{acl_natbib}

\appendix

\section{Supplementary Details and Results}
\label{sec:appendix}

\paragraph{Data and Metric} Table \ref{table-dataset} summaries the datasets and metrics we used, all datasets are using the public version from prior work \citet{sun2019lamol, chuang2020lifelong}\footnote{Datasets available at: \\ \url{https://github.com/chho33/LAMOL} \\ \url{https://github.com/voidism/L2KD}}. Note that some big datasets (WikiSQL, CNN/DailyMail, E2E NLG, RNNLG (laptop)) are reduced to a smaller size by random sampling due to data imbalance.

\begin{table}[h]
\begin{tabular}{llll}
\toprule
\textbf{Dataset}       & \textbf{Metric}                & \textbf{\# Train} & \textbf{\# Test} \\ \midrule
E2E NLG       & \multirow{5}{*}{ROUGE} & 6000     & 2000    \\
RNNLG(rest.)  &                        & 6228     & 1039    \\
RNNLG(hotel)  &                        & 6446     & 1075    \\
RNNLG(tv)     &                        & 8442     & 1407    \\
RNNLG(laptop) &                        & 7944     & 2649    \\ \hline
WikiSQL       & lfEM                   & 6525     & 15878   \\
CNN/DailyMail & ROUGE                  & 6604     & 2250    \\
MultiWOZ      & dsEM                   & 2536     & 1646    \\ \bottomrule
\end{tabular}
\caption{\label{table-dataset}
Dataset statistics and metrics. Note that ROUGE refers to the mean of ROUGE-1, ROUGE-2 and ROUGE-L, lfEM stands for exact match of logical forms, dsEM represents turn-based dialogue state exact match.
}
\end{table}

\paragraph{Task Sequences} In the scenario of \emph{CL on dissimilar tasks}, each task sequence also contains two or three similar natural language generation tasks, so the model cannot cheat by always adding new modules without detecting reusable ones.

\paragraph{Implementation Details}
We use GPT-2 \citep{radford2019language} in HugginceFace Transformers \citep{wolf2020transformers} as our backbone. We use the architecture from \citet{houlsby2019parameterefficient} in AdapterHub \citep{pfeiffer-etal-2020-adapterhub} with its default setting,
in which the reduce factor for bottle-neck architecture is $16$.
All experiments are conducted on NVIDIA RTX 2080 Ti with 11GB memory with a maximum batch size of 4. Training on one task sequence takes $5$ to $9$ hours. 

We use AdamW \citep{loshchilov2017decoupled} as our optimizer. We select learning rate from $\{1e-4, 1.75e-4, 3e-4\}$ and set the learning rate $lr=1.75e-4$ for all tasks except WikiSQL, and $lr=3e-4$ for WikiSQL. For decision stage, we train $6$ epochs to make decisions. For training stage, we select the best epoch number from $\{9, 12, 15\}$, and use $9$ for \emph{similar} scenario and $12$ for \emph{dissimilar} scenario. Weight initialization parameter $c$ is selected from $\{0.03, 0.05, 0.07\}$ for \emph{similar} scenario and  $\{0.12, 0.15, 0.17\}$ for \emph{dissimilar} scenario. Loss coefficient $\gamma$ is selected from $\{0.01, 0.05\}$, $\eta$ is set to $0.25$. Following \citet{sun2019lamol}, we use top-$k$ sampling where $k=20$ and set the pseudo-data sample rate to $0.2$. In our preliminary experiments, increasing the replay frequency can further alleviate forgetting. Thus, for those approaches using pseudo experience replay in this work, we set half of the training batches as pseudo-examples whenever learning a new task.

Note that the original design of Adapter+CL  \citep{madotto2020continual} uses perplexity to distinguish which task each testing example belongs to. In this work, we ignore that part and assume that the task-id of each testing example is given during inference for all baselines and our approach to ensure fair comparison.

\paragraph{Finetuning Results} We provide the results of finetuning GPT-2 \citep{radford2019language} and finetuning adapter \citep{houlsby2019parameterefficient} on all eight datasets in Table \ref{tab:finetune}. Since \citet{chuang2020lifelong} shows that the generation loss $L_{gen}$ could slightly increase the performance of finetuning on certain tasks, we also include the finetuning results after adding $L_{gen}$ loss.

Our results confirm that finetuning adapter can almost maintain the performance of finetuning the whole model. We also demonstrated that the performance of finetuning adapter could be improved by simply integrating $L_{gen}$ loss. This suggests that the performance of Adapter+CL could be naively improved by adding $L_{gen}$ to training loss. In that case, the average of mean score for Adapter+CL could be improved to 64.3 on \emph{similar} task sequences and 59.6 on $dissimilar$ task sequences, which are still significantly worse than our approach.

\begin{table}[h]
\centering
\setlength\tabcolsep{2pt}
\begin{tabular}{lccccc}
\toprule
\textbf{Method} & e2e & rest & hotel & tv & laptop\\
\midrule

GPT-2$_\text{finetune}\dagger$ & 48.8 & 64.0 & 65.4 & 70.8 & 73.0\\
GPT-2$_\text{finetune+gen} \dagger$ & 48.8 & 64.2 & 65.5 & 71.0 & 72.8 \\
Adapter$_\text{finetune}$ & 49.8 & 64.0 & 64.9 & 70.6 & 71.7 \\
Adapter$_\text{finetune+gen}$ & 49.9 & 64.3 & 65.1 & 70.6 & 71.8 \\
\midrule
\textbf{Method} & woz & cnn & wiki && \\
\midrule

GPT-2$_\text{finetune}\dagger$ & 84.8 & 25.5 & 63.1&&\\
GPT-2$_\text{finetune+gen} \dagger$ &82.2 & 25.9 & 63.7&&\\
Adapter$_\text{finetune}$ & 82.8 & 26.0 & 63.1&& \\
Adapter$_\text{finetune+gen}$ &83.5 & 26.0 & 63.8&& \\

\bottomrule
\end{tabular}
\caption{Finetuning results, $\dagger$ means we fetch numbers from \citet{chuang2020lifelong}}
\label{tab:finetune}
\end{table}

\paragraph{Results using Geometric Mean} While the mean of all tasks' performance score is always used \citep{sun2019lamol, mi2020continual, madotto2020continual} to represent the overall performance on several tasks, it could be largely influenced by the absolute change of one single number. In this work, we also leverage \emph{geometric mean} as an supplementary metric to measure the overall performance on different tasks, which provides another perspective to consider relative change during comparison.

Table \ref{table-res-geo} summarizes the final performance using geometric mean. We observed the same trend as in Table \ref{table-res1}, which demonstrates that our approach does improve the performance of baselines comprehensively on all tasks, not just in favor of absolute value increments on some tasks.

\begin{table*}[ht]
\centering
\small
\begin{tabular}{ll| p{1.3cm}<{\centering\arraybackslash} p{1.3cm}<{\centering\arraybackslash} p{1.3cm}<{\centering\arraybackslash}| p{1.3cm}<{\centering\arraybackslash} p{1.3cm} <{\centering\arraybackslash} p{1.3cm} <{\centering\arraybackslash} p{1.3cm}<{\centering\arraybackslash}}
\toprule
\multicolumn{2}{l|}{Methods} &
  \textbf{Finetune} &
  \textbf{EWC} &
  \textbf{LAMOL} &
  \textbf{\begin{tabular}[c]{@{}l@{}}Adapter\\ +CL\end{tabular}} &
  \textbf{\begin{tabular}[c]{@{}l@{}}Adapter\\ +Drop\end{tabular}} &
  \textbf{\begin{tabular}[c]{@{}l@{}}Adapter\\ +LAMOL\end{tabular}} &
  \textbf{Ours}  \\ \midrule
\multicolumn{2}{l|}{\begin{tabular}[c]{@{}l@{}}Pseudo \\ Experience Replay\end{tabular}} & \xmark & \xmark & \cmark & \xmark & \xmark & \cmark & \cmark \\ \midrule
\multirow{4}{*}{Similar Tasks}                            & \# 1                        &40.2 & 56.0 & 65.7  &  63.7&  63.4 & 65.4& \textbf{65.6}  \\
                                                          & \# 2                        &35.6 & 47.9& 66.3  &  63.7&   63.4&  65.5& \textbf{65.8}  \\
                                                          & \# 3                        &50.9  & 60.8& 66.0 &  63.7&   63.4& 64.9& \textbf{65.2}  \\
                                                          & \# 4                        &43.1  & 57.7 & 66.1 & 63.7&  63.4& 64.7& \textbf{65.2}   \\ \hline
\midrule
\multirow{4}{*}{Dissimilar Tasks}                         & \# 5                        & -- &  --&  54.3  &  53.7&   53.4&  47.8& \textbf{54.6} \\
                                                          & \# 6                        & -- & 24.0& 61.6  &  64.1&   63.6 &  61.2&  \textbf{65.0}  \\
                                                          & \# 7                        &16.8 & 36.1 & 53.4  &  53.5&   52.8&  51.3& \textbf{54.3}  \\
                                                          & \# 8                        &6.62 & 34.9 & 53.2 & 53.5&   52.8&  47.5& \textbf{54.8}  \\ \bottomrule
\end{tabular}
\caption{\label{table-res-geo}
Summary of final performance using geometric mean, where ``--`` denotes no valid geometric mean due to zero. We use two random seeds for each task sequence. Note that the final performance of Adapter+CL and Adapter+Drop is not affected by task ordering within the same group of tasks. For each sequence, we mark the best representation in \textbf{bold}, where LAMOL is not compared due to the difference in the order of magnitude of the learnable parameters.  
}
\end{table*}

\paragraph{Ablation Study}
Table \ref{table:ablation} summarizes the full details of ablation study conducted on sequence \#1 and \#8.

\paragraph{Detailed Final Performance}
Table \ref{table:detailed performance} provide the final performance of each task on every sequence for our approach and Adapter+LAMOL. For Adapter+CL, the final results are in Table \ref{tab:finetune}.

\begin{figure}[t]
\begin{center}
\includegraphics[width=0.9\linewidth]{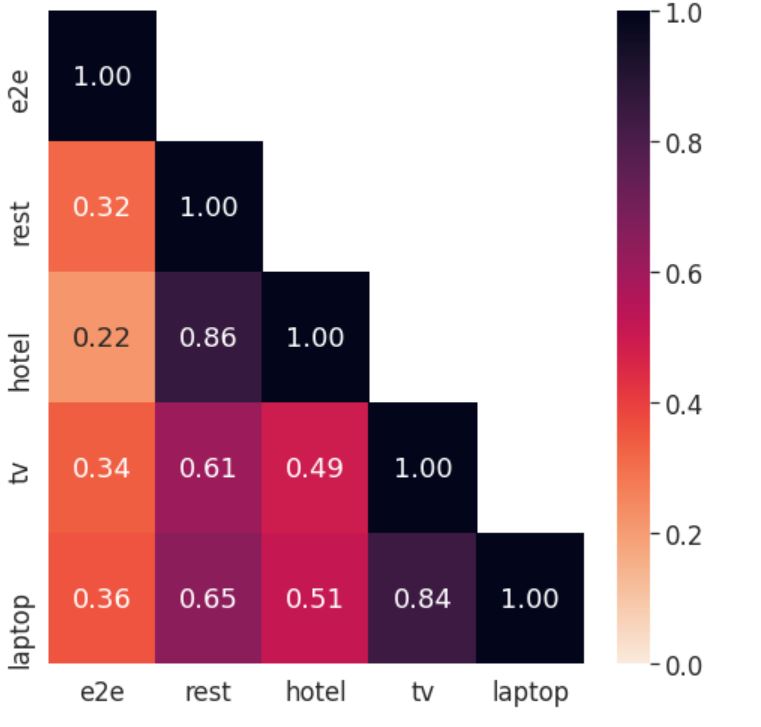}
\end{center}

\caption{Task Similarity calculated by the cosine similarity between each task's word frequency distribution.
}
\label{fig:task-sim}
\end{figure}

\paragraph{Task similarity}
Figure \ref{fig:task-sim} illustrates task similarity between five natural language generation tasks, which is calculated by the cosine similarity between each task's word frequency distribution.

% \begin{comment}
\begin{table*}[ht]
\centering
\small
\setlength\tabcolsep{3pt}

\begin{tabular}{l|ccccc|cc|ccccc|cc}
\toprule
\textbf{Method} & e2e & rest & hotel & tv & laptop & Avg & Avg L.P. & cnn & hotel & wiki & e2e & woz & Avg & Avg L.P. \\
\midrule

Ours & 51.7 & 66.7 & 67.7 & 72.4 & 71.9 & 66.1 & 2.24M & 27.8 & 65.3 & 62.9 & 51.7 & 83.3 & 58.2 & 6.49M\\
- Entropy loss & 52.1 & 67.1 & 67.6 & 72.3 & 71.5 & 66.1 & 2.54M & 27.8 & 64.8 & 62.6 & 49.8 & 82.9 & 57.6 & 6.49M\\
- Weight Ini & 49.6 & 64.7 & 64.8 & 70.4 & 71.3 & 64.2 & 7.09M & 26.7 & 64.7 & 64.6 & 49.9 & 82.4 & 57.7 & 8.65M\\
- Pseudo ER & 25.6 & 36.6 & 39.9 & 42.8 & 71.2 & 43.2 & 2.08M & 23.5 & 60.2 & 61.1 & 50.7 & 83.9 & 55.9 & 6.34M\\

\bottomrule
\end{tabular}
\caption{Ablation study on (i) entropy loss (ii) weight initialization (iii) pseudo experience replay. The left part includes results for sequence \#1 while the right part includes result for sequence \#8. Note that ``Avg`` refers to the mean of performance score on all tasks and ``Avg L.P.`` refers to the mean of learnable parameters.}
\label{table:ablation}
\end{table*}
% \end{comment}

\begin{table}[ht]
\centering
\small
\setlength\tabcolsep{3pt}
\begin{tabular}{l|ccccc|cc|ccccc|cc}
\toprule
\textbf{Method - \#1} & e2e & rest & hotel & tv & laptop & Avg \\
\midrule
Adap+LAMOL & 51.8 & 66.5 & 67.2 & 72.4 & 71.5 & 65.9 \\
Ours & 51.7 & 66.7 & 67.7 & 72.4 & 71.9 & 66.1 \\

\toprule
\textbf{Method - \#2} & laptop & tv & hotel & rest & e2e & Avg \\
\midrule
Adap+LAMOL & 74.8 & 75.2 & 65.9 & 66.0 & 49.3 & 66.2 \\
Ours & 64.7 & 74.5 & 51.5 & 73.5 & 49.7 & 66.5 \\

\toprule
\textbf{Method - \#3} & rest & tv & e2e & laptop & hotel & Avg \\
\midrule
Adap+LAMOL & 64.3 & 74.9 & 50.0 & 74.5 & 64.1 & 65.6 \\
Ours & 64.7 & 74.5 & 51.5 & 73.5 & 64.8 & 65.8 \\

\toprule
\textbf{Method - \#4} & hotel & e2e & rest & laptop & tv & Avg \\
\midrule
Adap+LAMOL & 66.4 & 50.9 & 65.8 & 73.0 & 70.0 & 65.2 \\
Ours & 66.4 & 51.3 & 66.2 & 74.2 & 70.6 & 65.7 \\

\toprule
\textbf{Method - \#5} & woz & cnn & e2e & rest & hotel & Avg \\
\midrule
Adap+LAMOL & 75.8 & 15.4 & 51.9 & 64.3 & 64.3 & 54.3 \\
Ours & 83.5 & 26.9 & 51.5 & 65.1 & 64.2 & 58.2 \\

\toprule
\textbf{Method - \#6} & e2e & wiki & hotel & woz & rest & Avg \\
\midrule
Adap+LAMOL & 53.4 & 47.9 & 64.6 & 80.4 & 64.7 & 62.2 \\
Ours & 50.9 & 64.3 & 65.1 & 84.1 & 64.8 & 65.9 \\

\toprule
\textbf{Method - \#7} & hotel & e2e & woz & wiki & cnn & Avg \\
\midrule
Adap+LAMOL & 66.0 & 48.5 & 77.5 & 55.4 & 25.8 & 54.6 \\
Ours & 67.0 & 50.9 & 83.5 & 64.1 & 25.9 & 58.3 \\

\toprule
\textbf{Method - \#8} & cnn & hotel & wiki & e2e & woz & Avg \\
\midrule
Adap+LAMOL & 16.5 & 65.2 & 52.5 & 51.4 & 83.4 & 53.8 \\
Ours & 27.8 & 65.3 & 62.9 & 51.7 & 83.3 & 58.2 \\

\bottomrule
\end{tabular}
\caption{Final Performance of each task on every sequence. Adap+LAMOL refers Adapter+LAMOL.}
\label{table:detailed performance}
\end{table}

\begin{table}[t]
\centering
\begin{tabular}{ccccc}
\toprule
\multicolumn{1}{c|}{\multirow{2}{*}{\textbf{Layer}}} &
  \multicolumn{2}{c|}{\textbf{Task A}} &
  \multicolumn{2}{c}{\textbf{Task B}} \\ \cline{2-5} 
\multicolumn{1}{c|}{} &
  \multicolumn{1}{c}{\textbf{O}} &
  \multicolumn{1}{c|}{\textbf{M}} &
  \multicolumn{1}{c}{\textbf{O}} &
  \textbf{M} \\ \midrule
1  & 59.6 & 72.5 & 95.1 & 92.5 \\
2  & 60.2 & 72.3 & 95.0 & 93.3 \\
3  & 60.1 & 71.3 & 95.1 & \emph{\underline{93.6}} \\
4  & 60.0 & 70.2 & 95.1 & 93.4 \\
5  & 60.0 & \emph{\underline{68.9}} & 95.2 & 91.3 \\
6  & 59.8 & 72.6 & 95.1 & 88.3 \\
7  & 60.0 & 71.2 & 95.0 & 86.2 \\
8  & 59.9 & 72.6 & 95.0 & 81.9 \\
9  & 59.6 & \textbf{76.7} & 95.0 & 83.8 \\
10 & 59.9 & 74.1 & 95.2 & 81.2 \\
11 & 59.9 & 74.5 & 95.0 & \textbf{80.3} \\
12 & 59.7 & 75.5 & 94.9 & 82.0 \\ \bottomrule
\end{tabular}
\caption{
Module Comparison: the effect of replacing the new module with the old module in different layers after sequentially learning Task A and B. Numbers in this table refer to the cosine similarity of word frequency distribution between \emph{the data of a specific task} and the $output$ \emph{generated from Task B's} $input$ (by original architecture - \textbf{O}, or modified architecture - \textbf{M}). We highlight the \textbf{most informative} layers and the \emph{\underline{least informative}} layers differently.
}
\label{tab:module-compare}
\end{table}

\section{Module Comparison}
\label{appendix: module comparison}

In order to demonstrate the compositional nature of our method, that is, each module contains different knowledge required for solving each task, we also study the performance difference to quantify the effect of reusing different modules.

\paragraph{Method} 
After training on task A, we specify a layer $k, k=1,2...12$ to add a new module for task B. Then we train the model on task B together with pseudo experience replay. After training on task B, we replace the new module with the old module from task A in layer $k$, and compare the performance difference on \emph{solving task B} between the modified architecture and the original architecture. On one hand, if the new added module contains specific knowledge of task B, then replacing it will result in the absence of corresponding feature in the generate $output$. On the other hand, if the old module contains specific knowledge of task A, then using it will result in some features of task A being generated in the $output$. 
% \xuezhi{this sentence is unclear}.
% further explained

\paragraph{Results}
Here we use laptop for task A and e2e for task B. We quantify the task knowledge contained in generated $output$ by calculating the cosine similarity of word frequency distribution between specific task's data and generated $output$. In Table~\ref{tab:module-compare}, we see that replacing the new module in layer 11 results in the most severe information loss of task B in the modified architecture, suggesting that the module in layer $11$ contains the most important information of word frequency for task B. In the same way, we conclude that module in layer $3$ contains the least important information of word frequency for task B. This is consistent with previous findings \citep{jawahar2019does} that bag-of-word information is mainly captured by higher transformer layers, while lower transformer layers capture surface and syntactic information.

Similarly, by analyzing the cosine similarity of word frequency distribution to task A, we find that the old module in layer $9$ contains the most important information of word frequency for task A and the old module in layer $5$ contains the least. While taking a closer look, we also find that modules in different layers contain information of different high-frequency words in task A. For example, module in layer $9, 10$ contains the most information of the word ``computing'', and ``laptop'', respectively, and module in layer $11$ contains more information of the word ``business'' than any other modules. This further demonstrates that different task-specific knowledge is contained in different modules from different layers, which results in different potential for reuse. By selectively reusing old modules to enable knowledge transfer and adding necessary modules to mitigate knowledge interference, our method derives a \textbf{compositional} architecture for every new task, as depicted in Figure \ref{fig:arch-growth}.

\end{document}